\documentclass[letterpaper, 10 j, conference]{ieeeconf}  

\IEEEoverridecommandlockouts                              

\usepackage[sorting=none]{biblatex}  
\usepackage{graphics} 
\usepackage{epsfig} 
\usepackage{times} 
\usepackage{amsfonts}
\usepackage{amsmath}

\usepackage{amssymb}
\usepackage[colorlinks, linkcolor=red, citecolor=blue]{hyperref}
\usepackage{algorithm}
\usepackage{algpseudocode}
\usepackage{epsfig}
\usepackage{graphicx}
\usepackage{caption}
\usepackage{subfigure}
\usepackage{mathrsfs}
\usepackage{array}
\usepackage{multirow}
\usepackage{multicol}
\usepackage{booktabs}
\usepackage{makecell}
\usepackage{xcolor}
\usepackage{amsfonts}
\usepackage{gensymb}
\usepackage{flushend}
\usepackage{steinmetz}
\usepackage{threeparttable}

\makeatletter
\newcommand{\tpmod}[1]{{\@displayfalse\pmod{#1}}}
\makeatother

\title{\LARGE \bf
DeepRING: Learning Roto-translation Invariant Representation for LiDAR based Place Recognition
}

\author{Sha Lu$^{1}$, Xuecheng Xu$^{1}$, Li Tang$^{2}$, Rong Xiong$^{1}$ and Yue Wang$^{1\dagger}$
\thanks{*This work was not supported by any organization}
\thanks{$^{1}$State Key Laboratory of Industrial Control and Technology, and the Institute of Cyber-Systems and Control, Zhejiang University, Hangzhou, 310058, China.} 
\thanks{$^{2}$Alibaba Group, Hangzhou, 310052, China.} 
\thanks{$^{\dagger}$Corresponding author wangyue@iipc.zju.edu.cn.}
}

\begin{document}

\maketitle
\thispagestyle{empty}
\pagestyle{empty}

\begin{abstract}


LiDAR based place recognition is popular for loop closure detection and re-localization. In recent years, deep learning brings improvements to place recognition by learnable feature extraction. However, these methods degenerate when the robot re-visits previous places with large perspective difference. To address the challenge, we propose DeepRING to learn the roto-translation invariant representation from LiDAR scan, so that robot visits the same place with different perspective can have similar representations. There are two keys in DeepRING: the feature is extracted from sinogram, and the feature is aggregated by magnitude spectrum. The two steps keeps the final representation with both discrimination and roto-translation invariance. Moreover, we state the place recognition as a one-shot learning problem with each place being a class, leveraging relation learning to build representation similarity. Substantial experiments are carried out on public datasets, validating the effectiveness of each proposed component, and showing that DeepRING outperforms the comparative methods, especially in dataset level generalization.

\end{abstract}

\section{INTRODUCTION}
Place recognition plays a significant role in autonomous driving applications. It retrieves the closest place from the past trajectory of robot for loop closure detection in SLAM systems to reduce the accumulated error. Because of the robustness to ever-changing environmental conditions, LiDAR sensors have been widely used for place recognition in recent years. As the field of view of LiDAR is wide, the LiDAR scans can have significant overlap when a robot revisits a previous place with different perspective. However, there still remains as a challenge for place recognition when a large perspective difference presents between the current scan and the scan taken at the previous place.

A popular way to deal with the challenge first occurs in handcrafted methods. By explicitly design the rotation invariant representation, scans taken by robot spinning in spot can keep the same. Therefore, the perspective difference visiting the same place can be suppressed, place can be determined by finding the most similar previous scan to the current one. Several methods have been proposed for achieving this property, including histogram, polar gram (PG), and principal component analysis \cite{rusu2010fast, rohling2015fast, kim2018scan, wang2020lidar, kim2021scan, li2021ssc}. More recently, translation invariance is also derived for the scan representation to address large perspective difference \cite{lu2022one}. However, the handcrafted feature extraction limits the discrimination of this line of works, most of whom employ simple features e.g. occupancy.

To improve the features, deep network is employed for feature learning from data \cite{uy2018pointnetvlad, Liu_2019_ICCV}. However, their networks do not inherently yield invariant representations, thus calling for data augmentation with artificially added rotation and translation to improve the robustness of scan representation again perspective change. Motivated by the representation design in handcrafted methods, in \cite{yin2018locnet, xu2021disco, chen2021overlapnet, li2022rinet}, the histogram, PG, and range image are inserted into neural networks to explicitly build rotation invariant representations. Together with deep learning, the discrimination can also be improved. Nevertheless, the rotation invariance loses when there is an obvious translation difference between the scans. It remains unclear how to keep deep features invariant when both rotation and translation differences present. 



\begin{figure*}[t]
	\centering
		\includegraphics[width=17cm]{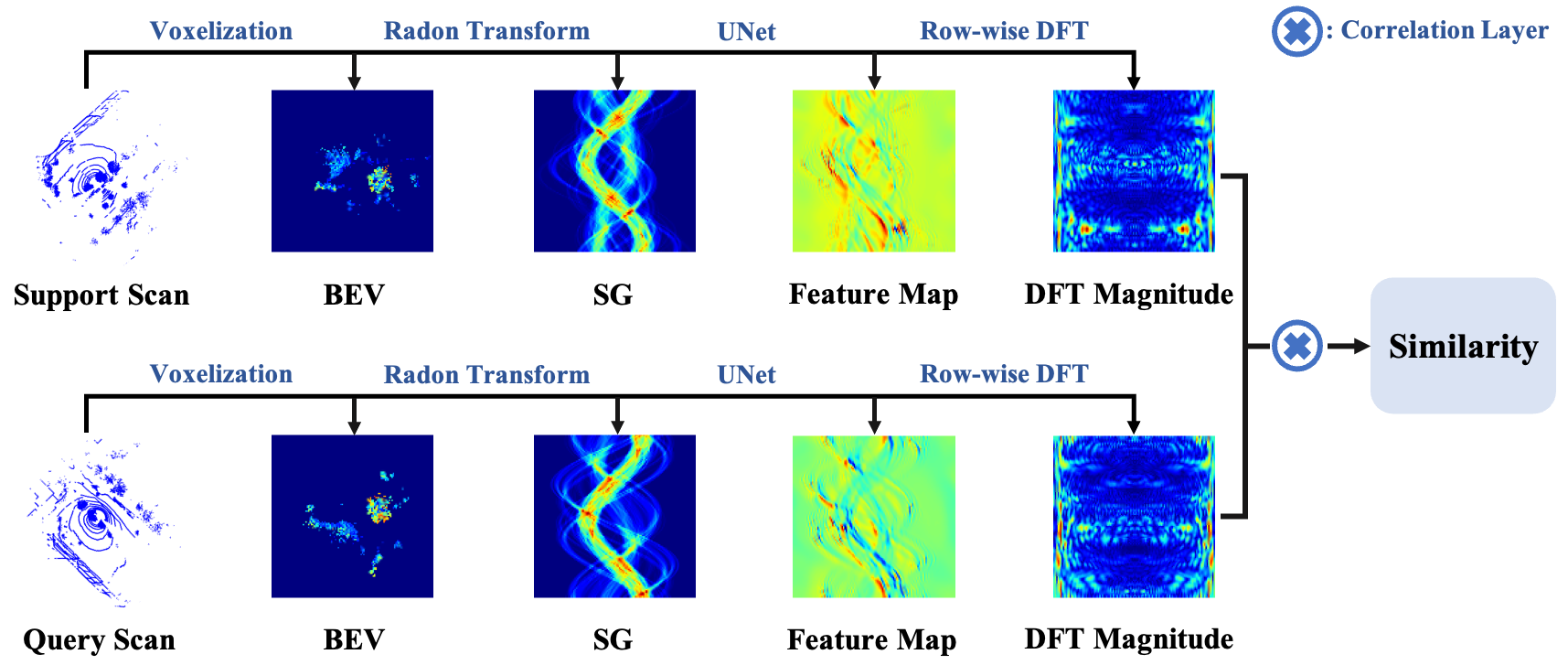}
	\caption{Overall framework of the proposed method DeepRING.}
	\label{fig:framework}
    \vspace{-0.5cm}
\end{figure*}

To address the problem, we propose a neural network architecture, named DeepRING, to learn \textit{roto-translation invariant} representation for a LiDAR scan by endowing the deep feature extraction with the RING architecture \cite{lu2022one}. As shown in Fig.~\ref{fig:framework}, we show that by formulating the LiDAR scan as sinogram (SG), taking it as input to the convolution network, and calculating the magnitude spectrum of the output, the resultant representation is roto-translation invariant, and can be learned from data for better discrimination in an end-to-end manner, bridging the advantages of two lines of works. Furthermore, we state the place recognition as a one-shot learning problem. Specifically, each place is regarded as a class, with the scan taken at that place as a shot, building a multi-class support set, while the current scan and the place form a query set. Such statement leverages the relation learning to replace the popular Siamese network and Euclidean triplet/quadruplet loss frequently used in place recognition. The experimental results validate the effectiveness of the proposed modules, showing that DeepRING outperforms the comparative methods, especially in data level generalization. In summary, the main contributions of our method comprise:
\begin{itemize}
\item
An end-to-end learning framework, DeepRING, to endow scan feature extraction with roto-translation invariant property, which tackles the problem of large perspective difference.
\item
Statement of place recognition as one-shot learning, to leverage relation learning for building better similarity. The efficient implementation saves the computation.
\item
Validation of the proposed method on two large-scale benchmark datasets, showing superior performance of DeepRING, especially for generalization, verifying the effectiveness of invariance design and one-shot learning. 
\end{itemize}


\section{RELATED WORKS}
In this section, we embark on related works review in terms of handcrafted methods and learning-based methods. In addition, we also introduce one-shot learning in brief.

\subsection{Handcrafted Methods}
Fast Histogram \cite{rohling2015fast} leverages the range of 3D points and encode it into histogram as the global descriptor. M2DP \cite{he2016m2dp} projects the LiDAR scan into multiple 2D planes and generate the global descriptor that is robust to rotation change via PCA. Scan Context series \cite{kim2018scan, kim2021scan} utilize an egocentric spatial descriptor encoded by the maximum height of points. Likewise, LiDAR-Iris \cite{wang2020lidar} extracts the LiDAR-Iris binary image and transforms it into frequency domain to achieve rotation invariance. RING \cite{lu2022one} proposes sinogram to represent point clouds for both place recognition and pose estimation. In these methods, the similarity computation generally calls for an exhaustive search. Moreover, the feature extraction in these methods limits the discrimination of the representation.   

\subsection{Learning-based Methods}
PointNetVLAD \cite{uy2018pointnetvlad} extracts features from the raw 3D point cloud with PointNet \cite{qi2017pointnet} and aggregates them to global descriptor with NetVLAD \cite{arandjelovic2016netvlad}. LPD-Net \cite{Liu_2019_ICCV} proposes a graph-based neighborhood module to aggregate the extracted adaptive local features, which enhances the place description of the global descriptor especially for large-scale environments. Locus \cite{vidanapathirana2021locus} aggregates the topological relationship with temporal information of point clouds to improve the place description ability. Expect for global descriptors, some works \cite{du2020dh3d, komorowski2021egonn, cattaneo2022lcdnet} learn both global and local features of LiDAR scans to address a 6DoF localization problem. Without explicit invariant representation, these methods achieve robustness against perspective difference by data augmentation. LocNet \cite{yin2018locnet} encodes the range histogram as fingerprint and learns a semi-learning neural network to achieve rotation invariance. OverlapNet \cite{chen2021overlapnet} exploits multiple cues from scans and predicts the overlap together with the relative yaw angle between two scans using a Siamese network. DiSCO \cite{xu2021disco} converts the point cloud in the cylindrical coordinate, extracts features using a encoder-decoder network and transforms these features to frequency domain to reach rotation invariance. RINet \cite{li2022rinet} further exploits the stage of inserting feature extraction to keep the rotation invariance. However, the rotation invariance in these methods are sensitive to translation difference. A larger translation may degenerate their performances.

\subsection{One-shot Learning}
With the progress of deep learning in the past decades, learning based methods have presented excellent performance. A typical scenario is that the class in test phase are all seen in training phase, and in each class there are lots of samples. However, this is not true in tasks like face recognition, or place recognition, where each class only has few samples, and the class in test phase are all unseen. To deal with this problem, one-shot learning is proposed, a typical method is distance metric learning based via "learning to compare with distance metrics" \cite{chen2019closer}. For instance, Matching Network \cite{vinyals2016matching} employs cosine similarity as the distance metric. Prototypical Network \cite{snell2017prototypical} utilizes Euclidean distance in the embedding space to compare different classes. Relation Network \cite{sung2018learning} designs a CNN to learn a distance metric to compare the relation of images. Inspired by the works in this direction, we state the place recognition problem as one-shot learning to leverage fruitful relation learning methods to build the similarity between scans. 

\section{METHODOLOGY}

We propose a one-shot learning framework based on sinogram (SG) representation, named DeepRING, to construct roto-translation invariance for robust place recognition. 

\subsection{Rotation Equivariant Representation}
\label{rer}
\textbf{Sinogram:}
In this subsection, we convert a LiDAR scan to a sinogram representation which is visualized in Fig.~\ref{fig:framework}. Given a raw 3D scan, we first remove the uninformative points of ground plane and quantize the preprocessed point cloud into a finite number of pillars. Taking advantage of the occupancy information in each pillar, we encode the 3D point cloud into a bird-eye view (BEV) image, denoted as $f(x,y)$. After that, we apply Radon transform to the BEV image, yielding a resultant sinogram (SG) denoted as $R_{f}(\theta, \tau)$. The mathematical formula of Radon transform along a couple of parallel lines is 
\begin{equation}
\begin{split}
&\mathcal{R}_{f}(\theta, \tau)=\int_{L: \; x\cos\theta + y\sin\theta = \tau} f(x,y) \mathrm{d}x \mathrm{d}y \\
&=\int_{-\infty}^{\infty}\int_{-\infty}^{\infty} f(x,y) \delta(\tau-x\cos\theta-y\sin\theta) \mathrm{d}x \mathrm{d}y
\end{split}
\end{equation}
where $L$ represents the integrated line parameterized as $x\cos\theta + y\sin\theta = \tau$, $\theta \in [0, 2\pi)$ is the angle between $L$ and the $y$ axis, and $\tau \in (-\infty, \infty)$ is the perpendicular distance from the origin to $L$.

Radon sinogram transforms a rotation in the BEV image space (corresponding to the 3D Cartesian space) to an equivariant circular shift along $\theta$ axis in the radon image space, which is also rotation equivariant. Specifically, applying the rotation by an angle $\alpha$ to the raw 3D scan, the resultant sinogram shifts along $\theta$ axis by the distance $\alpha$, which can be written as
\begin{equation}
\mathcal{R}_{f}(\theta, \tau) \stackrel{\alpha}{\longrightarrow}  \mathcal{R}_{f}(\theta + \alpha, \tau)	
\label{rot}
\end{equation}

Aside from rotation transformation, a translation by a vector $\vec{d} = (\Delta{x},\Delta{y})^T$ reflects a linear shift in the variable $\tau$ of sinogram, equal to the projected length of vector $\vec{d}$ onto line $xcos{\theta} + ysin{\theta} = \tau$, namely, $\Delta{\tau} = (\cos\theta,\sin\theta)\vec{d} = \Delta{x}cos{\theta} + \Delta{y}sin{\theta}$. The corresponding relationship is
\begin{equation}
\mathcal{R}_{f}(\theta,\tau) \stackrel{\vec{d}}{\longrightarrow} \mathcal{R}_{f}(\theta, \tau - \Delta{\tau})	
\label{trans}
\end{equation}

\textbf{Other Representations:}
In terms of intermediate representations, polar transform and spherical projection are widely used to convert a point cloud from the 3D Cartesian space to the 2D image space. Similar to SG, a rotation in the 3D space is projected onto a cyclic shift along the rotation-related dimension in the image space, arriving at rotation equivariance. However, a translation causes a nonlinear shift along the rotation-related dimension of PG and range image, resulting in a changing scale of these two representations. The nonlinear shift along the axises of PG is mathematically written as
\begin{align}
\begin{split}
&\mathcal{P}_{f}(r,\theta) \stackrel{\vec{d}}{\longrightarrow} \mathcal{P}_{f}(r', \theta')\\ 
&r' = \sqrt{(r\cos\theta-\Delta{x})^2+(r\sin\theta-\Delta{y})^2}\\
&\theta' = \arctan\frac{r\sin\theta-\Delta{y}}{r\cos\theta-\Delta{x}}
\label{nonlinear_shift}
\end{split}
\end{align}
where $\mathcal{P}_{f}(r,\theta)$ is the result after polar transform, i.e. PG. By comparing Eq.~\ref{nonlinear_shift} with Eq.~\ref{trans}, we can see that the translation variance severely corrupts the rotation equivariance property of PG. Same for range image, the translation influences the scale of the range image, which is also harmful for the rotation equivariance.


Under this circumstance, it is difficult to eliminate translation variance, which simultaneously affects the strict rotation equivariance of these representations in practice.

\subsection{Feature Extraction Network}
After the intermediate rotation equivariant representation construction, we utilize a feature extraction network to generate a more discriminative place representation. In order to maintain the rotation equivariance, relative operations are required, which corresponds to translation equivariant operations in our case. Standard convolutions are equivariant to translations so that they are successfully applied in many image tasks. Nevertheless, the circular shift resulted from rotation is circularly bounded in $[0, 2\pi)$. Consequently, conventional linear convolutions can not guarantee strict rotation equivariance because they fail to deal with data in the border of the image. To address this problem, we choose circular convolutions which extract consistent features along the upper and lower boundaries of the image, formulated as
\begin{equation}
\begin{split}
\label{circonv}
&E_{f}[i,j] = ({R}_{f} \circledast K)[i,j] \\
&=\sum_{m=0}^{M-1}\sum_{n=0}^{N-1}({R}_{f}[(i-m) \%{M},(j-n) \%{N}]K[m,n])
\end{split}
\end{equation}
where $\circledast$ denotes the circular convolution operation, $\%$ denotes the mod operation, $E_{f}$ is the output feature map, ${R}_{f} \in \mathbb{R}^{M \times N}$ is the input sinogram, and $K \in \mathbb{R}^{M \times N}$ is the convolution kernel. 
Note that $E_{f}$ is translation equivariant to $R_{f}$ resulting from the translation equivariant property of circular convolutions. Combining Eq.~\ref{trans} and Eq.~\ref{circonv}, we can arrive at
\begin{equation}
E^{d}_{f}(\theta_i, \tau) = E_{f}(\theta_i, \tau-\Delta{\tau})
\end{equation}
where ${E}^{d}_{f}$ represents the output feature map after translation on the input scan by $\vec{d}$.

After the last circular convolution layer of the feature extraction network, the output $E_{f}$ is passed through row-wise DFT to eliminate translation changes. Referring to the translation invariant property of DFT, the translation invariance is arrived by taking the magnitude of the frequency spectrum. Denote the resultant magnitude spectrum as ${M}_{f}$, then we have 
\begin{equation}
\begin{split}
{M}^{d}_{f}(\theta_i, \omega) &= |\mathcal{F}(E^{d}_{f}(\theta_i, \tau))| \\
&= |\mathcal{F}(E_{f}(\theta_i, \tau-\Delta{\tau}))| \\
&= |\mathcal{F}(E_{f}(\theta_i, \tau))||e^{-j 2\pi \omega \Delta{\tau}}| \\
&= |\mathcal{F}(E_{f}(\theta_i, \tau))| = {M}_{f}(\theta_i, \omega)
\end{split}
\end{equation}
where $|\cdot|$ is the magnitude operation, $\mathcal{F}(\cdot)$ is the 1D DFT operation, ${M}^{d}_{f}$ is the DFT magnitude of ${E}^{d}_{f}$, $\omega$ is the discrete sampled frequency and $j$ is the imaginary unit. With translation invariance, the impact of translation disturbance on rotation equivariance is further alleviated. In this paper, we utilize UNet \cite{ronneberger2015u} equipped with circular convolutions and DFT as the feature extraction network, as depicted in Fig. \ref{fig:framework}.

\begin{figure}[t]
	\centering
		\includegraphics[width=8.6cm]{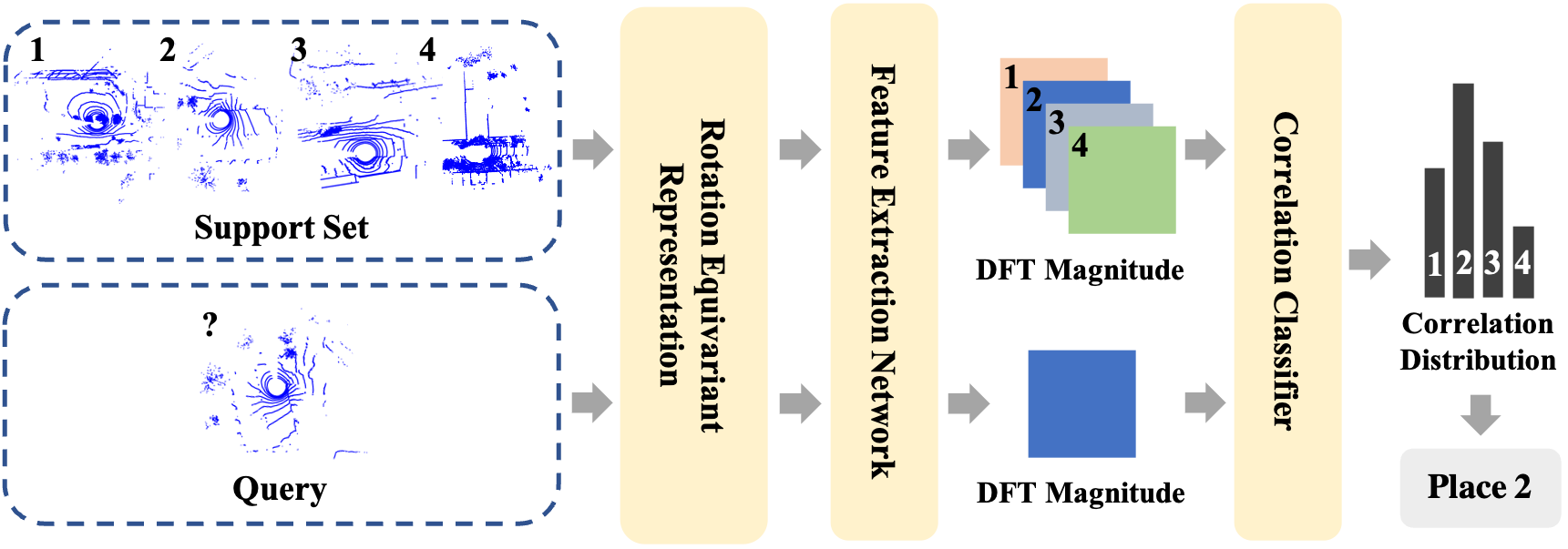}
	\caption{One-shot learning algorithm for place recognition.}
	\label{fig:oneshot}
    \vspace{-0.5cm}
\end{figure}

\subsection{Statement as One-shot Learning}

Place recognition aims to recognize the same place from all places in the database, which can be seen as a multi-way one-shot learning problem. Each place can be regarded as a class, and the scan taken from this place is a shot. Therefore, we tailor the one-shot learning algorithm for place recognition, as illustrated in Fig. \ref{fig:oneshot}. In the training stage, the support set and query set are randomly chosen from the map database. In the test stage, the query is the scan taken from the current place. The similarity functions in one-shot learning pipelines vary with different feature representations. Cosine and Euclidean distance are frequently used to calculate similarity for common features. However, these distance metrics do not maintain the internal relations between the feature maps learned from equivariant SG representations.

To better leverage the relations between SG representations, we propose a correlation-based distance metric to compare the similarity between places. The circular cross-correlation layer is employed after the feature extraction network to compute the correlation between support and query scans. Based on rotation equivariant and translation invariant feature maps, the correlation distance metric is symmetric and roto-translation invariant, which is defined as
\begin{equation}
   \mathcal{C}({M}_{f_1}, {M}_{f_2}) = \max_{\alpha} \sum^{}_{\theta} \sum^{}_{\omega} {M}_{f_1}(\theta + \alpha, \omega) {M}_{f_2}(\theta, \omega)
   \label{corr}
\end{equation}
where $\mathcal{C}({M}_{f_1}, {M}_{f_2})$ denotes the resultant correlation value between ${M}_{f_1}$ and ${M}_{f_2}$, and $\alpha$ corresponds to the shift at the best alignment.
For each query, we can obtain a correlation vector $\mathfrak{C} \triangleq \{\mathcal{C}({M}_{f_Q}, {M}_{f_S}^{(i)})\}$ consisting of the correlation values between the query and each shot in the support set.

After that, we normalize the correlation values in $\mathfrak{C}$ with a softmax layer, given to the classifier to predict the class. 
\begin{equation}
\tilde{\mathfrak{C}} = Softmax(W \mathfrak{C} + b) \quad (W, b \in \mathbb{R})
\end{equation}
where $\tilde{\mathfrak{C}} \in [0, 1]$ indicates the final correlation vector after softmax normalization, which is shown in Fig.~\ref{fig:oneshot}.

Instead of using BCE loss for a binary classifier like \cite{li2022rinet, koch2015siamese}, we utilize cross-entropy loss considering place recognition as a multi-class classification problem.


\subsection{Implementation Details}
In the training phase, we train the model for 20 epochs, each of which incorporates 60 episodes. We use Adam \cite{kingma2015ba} optimizer with the initial learning rate of $10^{-3}$ and decay of $10^{-4}$. To schedule the learning rate, we use MultiStepLR with gamma of 0.1 and milestones at 5 and 12 epoch. Following the episodes construction proposed in \cite{vinyals2016matching}, we randomly select 24 classes from the training data and sample 7 examples including 1 shot and 6 queries within these classes to form support set $S$ and query set $Q$ respectively in each episode. To distinguish different classes in place recognition problem, we consider that scans whose positions are within $10m$ from each other belong to the same class and scans whose positions are beyond $20m$ from each other belong to different classes.

\section{DATASETS}
\begin{figure}[t]
	\centering
    \subfigure[Disjoint NCLT Dataset]{
		\includegraphics[width=3.5cm]{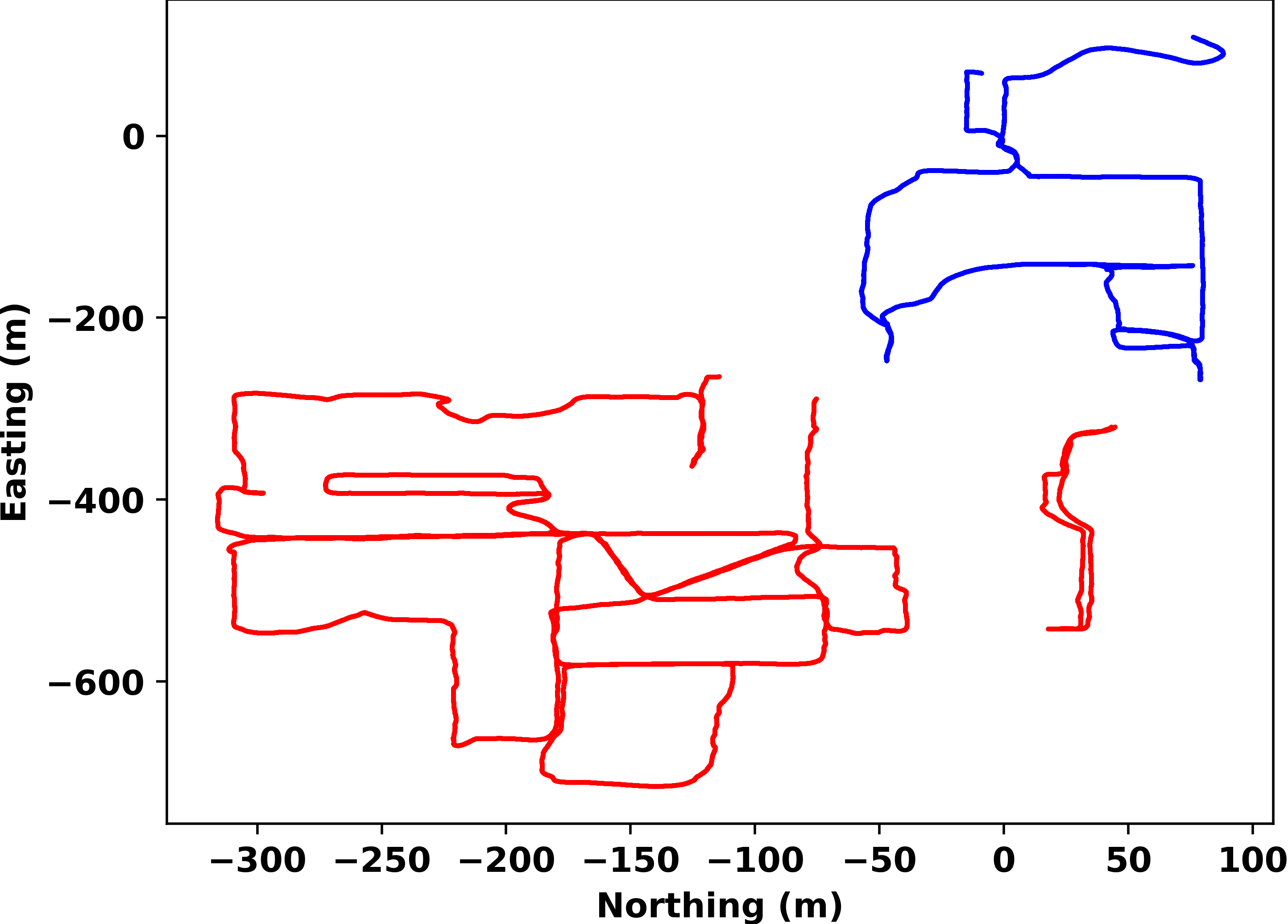}}
    \subfigure[Disjoint MulRan Dataset]{
		\includegraphics[width=3.5cm]{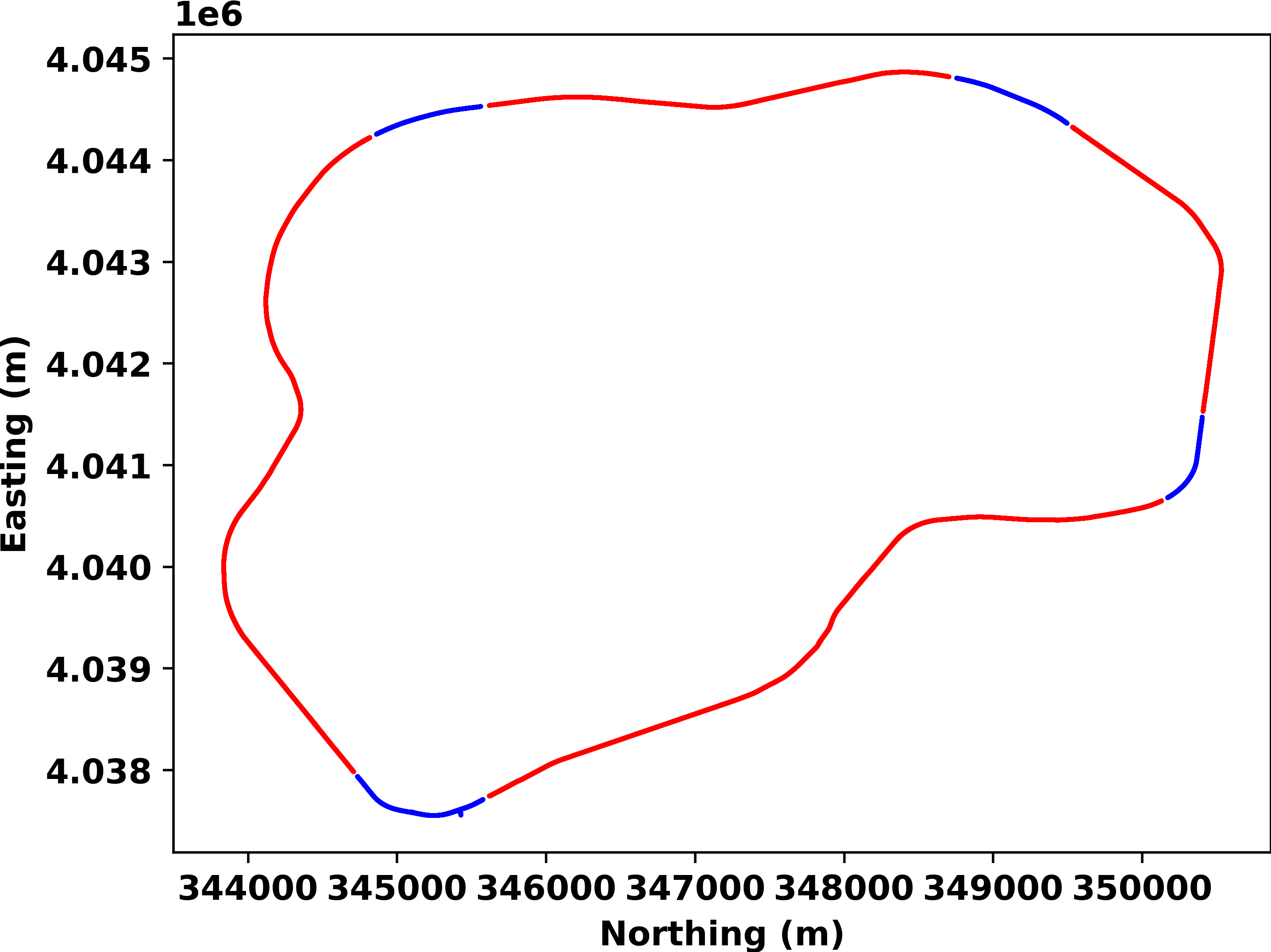}}
	\caption{Disjoint sessions for training (red) and test (blue).}
	\label{fig:split}
    \vspace{-0.5cm}
\end{figure}

We evaluate our method on disjoint regions of the sessions from two benchmark datasets NCLT \cite{carlevaris2016university} and MulRan \cite{kim2020mulran}, and the detailed splits of training and test are shown in Fig.~\ref{fig:split}. 
We evaluate the method with sparse data: the sampling distance of database and query trajectory are $20m$ and $5m$. 
\subsection{NCLT Dataset}
The NCLT Dataset contains data of 27 mapping sessions over 15 months collected by a Segway robot. It includes a variety of environmental changes covering seasonal, temporal, and structural changes. Velodyne HDL-32E 3D LiDAR is used for point clouds acquisition. Among these sessions, we choose ``2012-02-04'' and ``2012-03-17'' sessions for model training and test, where ``2012-02-04'' serves as database session and ``2012-03-17'' serves as query session.

\subsection{MulRan Dataset}
The MulRan Dataset is a multi-modal range dataset for large-scale place recognition evaluation. It acquires scan data using Ouster OS1-64 LiDAR under various environments in South Korea. In the experiments, we select Sejong01 as database session and Sejong02 trajectories for training and test, where Sejong01 serves as database session and Sejong02 serves as query session. 


\subsection{Evaluation Metric}
To evaluate place recognition performance of the proposed method comprehensively, we utilize four standard metrics: \textit{Recall@1} reports the percentage of correctly recognized place using top 1 candidate; \textit{F1 Score} takes the harmonic mean of precision and recall, combining them into a single metric; \textit{Precision-Recall Curve} represents the performance of a classifier across a number of thresholds; \textit{AUC} computes the area under the \textit{Precision-Recall Curve}.


\section{EXPERIMENTAL RESULTS}
In this section, based on the datasets and metric above, we conduct a case study to visualize the roto-translation invariant representation, a comparative study to evaluate the performance and generalization, and several ablation studies to validate each proposed component.

\subsection{Case Study}
\label{case}
\begin{figure}[t]
	\centering
    \includegraphics[width=8.4cm]{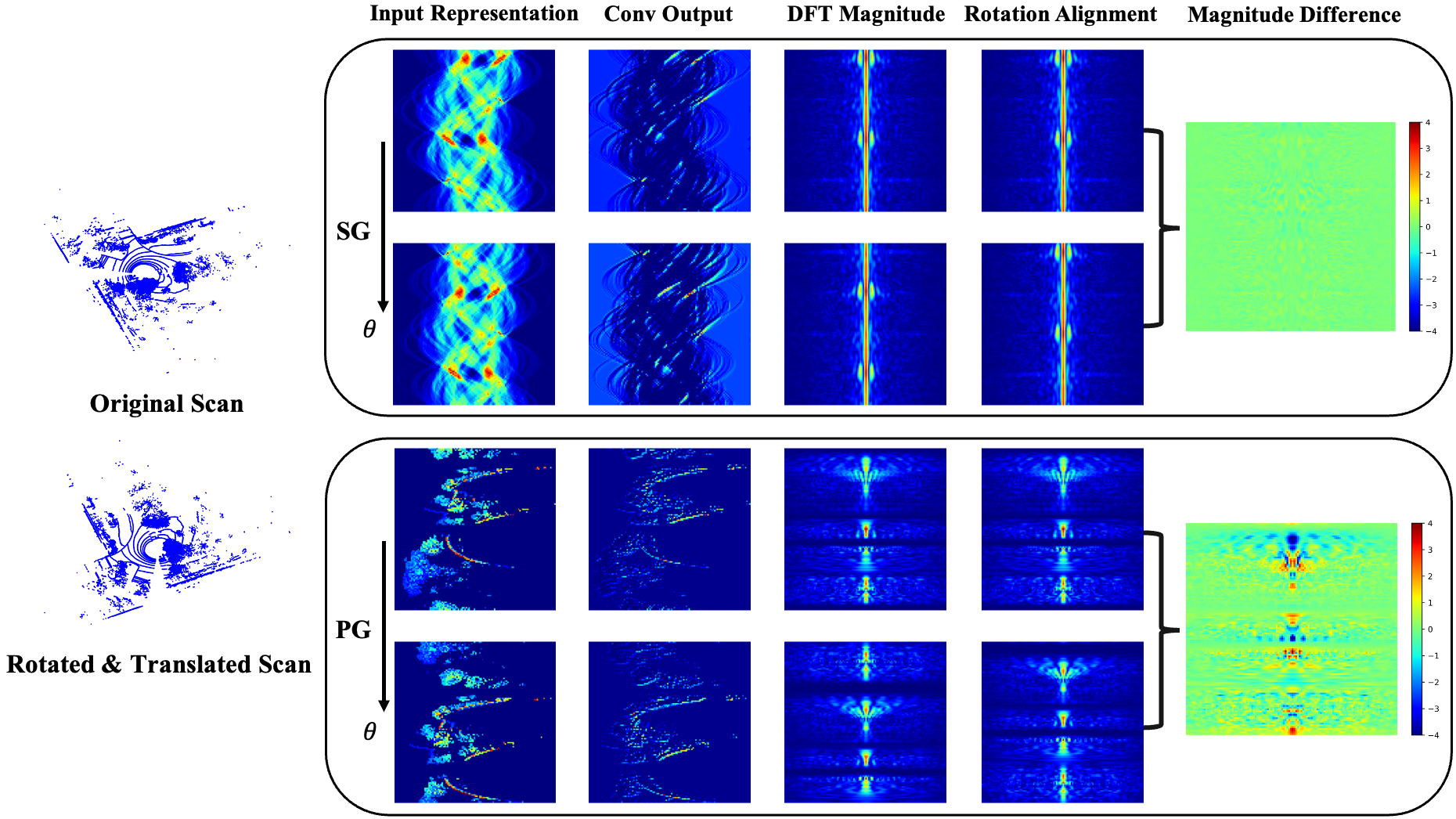}
	\caption{Case study for roto-translation invariance achievement based on SG and PG, where the vertical arrow represents the rotation related $\theta$ axis of SG and PG.}
	\label{fig:case}
    \vspace{-0.5cm}
\end{figure}

We first design a case study to validate that SG is a better representation than PG for roto-translation invariance as mentioned in Sec.~\ref{rer}. We apply a random transformation comprising rotation and translation to generate a pair of point clouds, and then convert them to SG and PG representations respectively. Then we employ a circular convolution layer followed by row-wise DFT as a simplified implementation of feature extraction network on these two representations. Finally, we manually align the vertical axis of DFT magnitudes from SG and PG to eliminate the effect of rotation. In theory, if the representation is rotation equivariant and translation invariant, the difference between two scan representations should be zero. As shown in Fig.~\ref{fig:case}, SG based representation leads to obviously smaller difference, validating its invariance.

\subsection{Comparison with State-of-the-Art}
We compare our method with the state-of-art handcrafted and learning-based methods: Scan Context \cite{kim2018scan}, RING \cite{lu2022one}, PointNetVLAD \cite{uy2018pointnetvlad}, DiSCO \cite{xu2021disco} and EgoNN \cite{komorowski2021egonn}. For all approaches, we leverage the same data preprocessing for fair comparison. We utilize the same resolution of $120 \times 120$ for quantization in Scan Context, RING and our method to compare the performance of different representations. For EgoNN, we use the pre-trained model released publicly for evaluation on MulRan dataset and retrain the model on NCLT dataset for evaluation on NCLT dataset. We retrain PointNetVLAD and DiSCO on both datasets. Additionally, all parameters and details keep the same as the original papers. 

\begin{figure}[t]
	\centering
    \includegraphics[width=8.6cm]{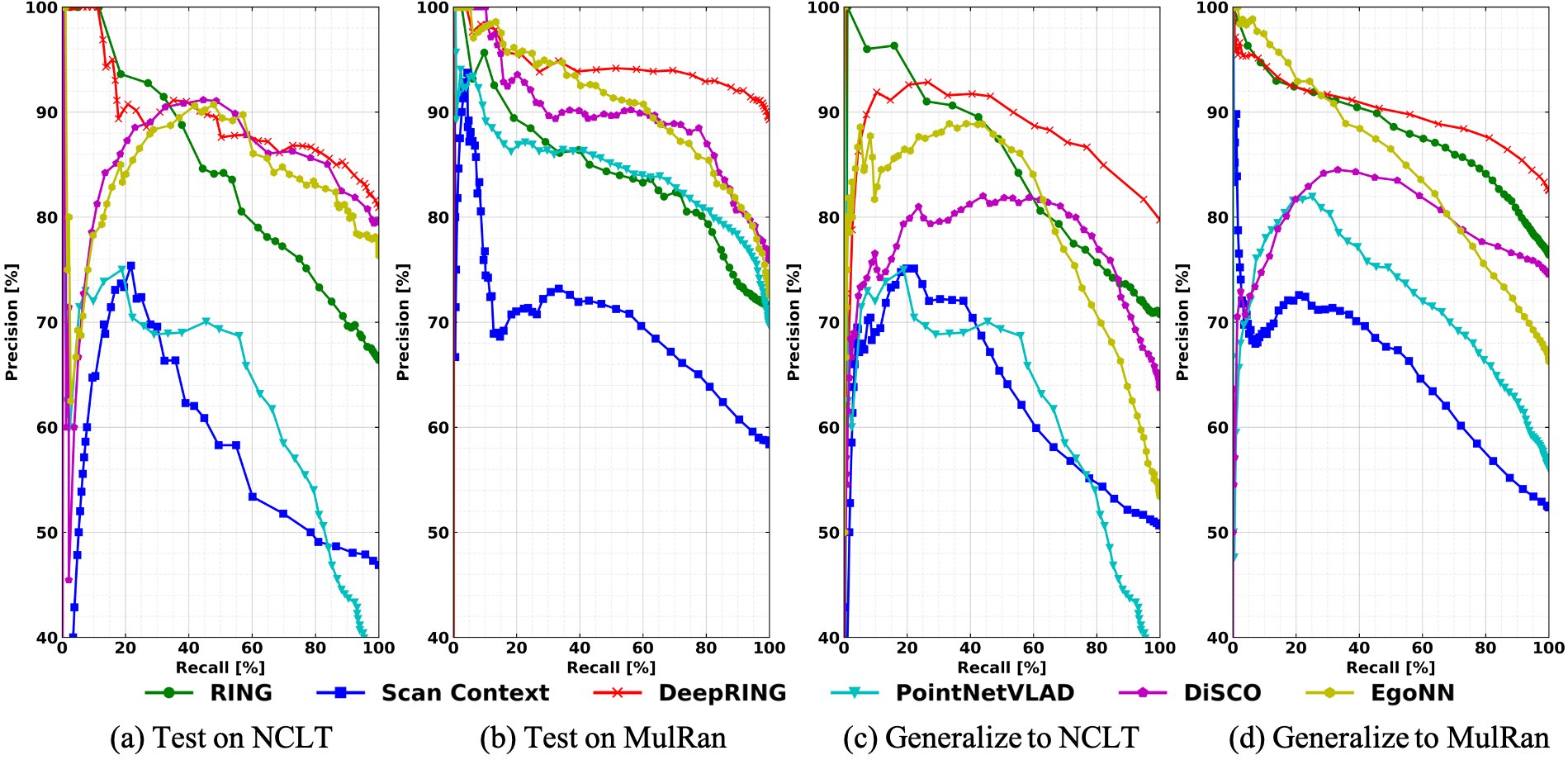}
	\caption{Precision-Recall Curve of comparative methods.}
	\label{fig:prcurve}
    \vspace{-0.5cm}
\end{figure}

\textbf{Evaluation Performance:} Tab.~\ref{evaluation} and Fig.~\ref{fig:prcurve} depict the place recognition performance of all methods on the test trajectory of NCLT and MulRan datasets. In terms of two handcrafted methods, Scan Context reaches rotation invariance by brute-force matching and RING is roto-translation invariant taking advantage of the translation invariant property of DFT, which explains the advantage over PointNetVLAD. But limited by the discrimination of handcrafted features, the performances of these two methods are inferior than DiSCO that combines rotation invariance and feature learning. EgoNN shows approximate performance with DiSCO owing to its more complex feature extraction, as well as approximated rotation invariant representation. With both roto-translation invariant representation design and the learnable feature extraction, DeepRING outperforms the other learning based methods and handcrafted methods on the test datasets.

\textbf{Generalization Performance:}
To evaluate the cross-domain generalization, we train the deep learning models on MulRan (NCLT) dataset and evaluate them on NCLT (MulRan) dataset, with the results depicted in Tab.~\ref{generalization} and Fig.~\ref{fig:prcurve}. In a learning-free manner, Scan Context and RING show stable performance on both two datasets. In contrast, the learning based methods have degraded performance in the generalized domain, especially for PointNetVLAD due to the less inductive bias on invariance. Compared with PointNetVLAD, EgoNN refers to PG representation, accounting for better generalization to unseen places. DiSCO exceeds EgoNN by the explicit rotation invariant representation. Benefited from the explicit roto-translation invariant representation and one-shot learning, DeepRING shows the best generalization in new domain.

\subsection{Ablation Study}
To investigate the specific contribution of representation, aggregation and loss modules, we design multiple ablation studies on NCLT dataset and show the results in Tab. \ref{total ablation}. 

\textbf{Representation:}
We compare the performance of two rotation equivariant representations, PG and SG, in terms of place recognition. As we can see, SG based model outperforms PG based model, which is consistent with case study in Sec.~\ref{case}. The underlying reason is that rotation and translation transformations are decoupled in SG representation, rotation is not sensitive to translation difference. In contrast, translation variance deteriorates the rotation equivariance property as analyzed in Sec.~\ref{rer}.

\textbf{Aggregation:}
After feature extraction, feature aggregation is employed to build scan representation. In the proposed framework, we employ row-wise DFT to aggregate the features into a global translation invariant representation. In addition, operations like global max pooling and average pooling can also arrive at translation invariance, which are widely used to aggregate features in many works. Therefore, we compare DFT against these two aggregation approaches according to the results in Tab.~\ref{total ablation}. Global average pooling with \textit{Recall@1} of 0.639 outperforms global max pooling with \textit{Recall@1} of 0.534, because of the non-linearity caused by max operation. Since the global pooling operation squeezes the translation dimension while DFT following by magnitude operation does not change the size of feature maps, we increase the kernel number of the last convolution layer to yield a multi-channel feature map so that the resultant representation after pooling keeps the same size as that after DFT. Compared to the improved global average pooling with \textit{Recall@1} of 0.743, the DFT based method with \textit{Recall@1} of 0.859 still improves the performance substantially. Therefore, we consider that DFT is learning-free, thus making it a better aggregation method to build data independent translation invariance, while keeping discrimination.

\textbf{Loss:}
The mainstream metric learning based methods take triplet loss or quadruplet loss as the loss function. For comparison, we replace the cross-entropy loss of our method by triplet loss and evaluate the final place recognition performance. The triplet loss based model results in inferior performance, with \textit{Recall@1} decreasing from 0.859 to 0.644. The triplet loss is usually accompanied by Euclidean distance which does not support the rotation equivariance of DFT magnitude, but the cross-entropy loss of the proposed method actually finds the relative rotation and similarity at the same time. For a more fair comparison, we further employ an extra column-wise DFT after DFT magnitude to eliminate the effect of rotation. As a result, \textit{Recall@1} increases to 0.747, which is still surpassed by our method. It demonstrates that the cross-entropy loss based on correlation distance is more suitable for roto-translation invariant feature learning.

\begin{table}[t]
\renewcommand\arraystretch{1.2}
\centering
\caption{Place Recognition Performance Comparison}
\label{evaluation}
\begin{threeparttable}
\begin{tabular}{clccc}
\toprule[1pt]
\textbf{Dataset} & \textbf{Approach} & \textbf{Recall@1} & \textbf{F1 Score} & \textbf{AUC} \\ \hline
\multirow{6}{*}{NCLT} & Scan Context & 0.469 & 0.638 & 0.570 \\ 
& RING & 0.664 & 0.798 & 0.839 \\ 
& PointNetVLAD & 0.508 & 0.539 & 0.627 \\ 
& DiSCO & \underline{0.793} & \underline{0.887} & \underline{0.849} \\ 
& EgoNN & 0.763 & 0.866 & 0.842 \\ 
& \textbf{DeepRING (ours)} & \textbf{0.859} & \textbf{0.894} & \textbf{0.893} \\ \hline
\multirow{6}{*}{MulRan} & Scan Context & 0.584 & 0.737 & 0.698 \\ 
& RING & 0.713 & 0.833 & 0.843 \\ 
& PointNetVLAD & 0.697 & 0.821 & 0.842 \\ 
& DiSCO & \underline{0.743} & \underline{0.863} & 0.897 \\ 
& EgoNN & 0.726 & 0.841 & \underline{0.906} \\ 
& \textbf{DeepRING (ours)} & \textbf{0.893} & \textbf{0.943} & \textbf{0.943} \\
\bottomrule[1pt]
\end{tabular}
\begin{tablenotes}
        \footnotesize
        \item[*] The compared methods are evaluated on the test trajectory.
      \end{tablenotes}
\end{threeparttable}
\vspace{-0.6cm}
\end{table}

\begin{table}[t]
\renewcommand\arraystretch{1.2}
\centering
\caption{Cross-domain Generalization Evaluation}
\label{generalization}
\begin{threeparttable}
\begin{tabular}{clccc}
\toprule[1pt]
\textbf{Dataset} & \textbf{Approach} & \textbf{Recall@1} & \textbf{F1 Score} & \textbf{AUC} \\ \hline
\multirow{6}{*}{NCLT} & Scan Context & 0.507 & 0.673 & 0.629 \\ 
& RING & \underline{0.708} & \underline{0.829} & \underline{0.847} \\ 
& PointNetVLAD & 0.369 & 0.539 & 0.626 \\ 
& DiSCO & 0.604 & 0.779 & 0.773 \\ 
& EgoNN & 0.562 & 0.696 & 0.799 \\ 
& \textbf{DeepRING (ours)} & \textbf{0.797} & \textbf{0.887} & \textbf{0.881} \\ \hline
\multirow{6}{*}{MulRan} & Scan Context & 0.524 & 0.688 & 0.653 \\ 
& RING & \underline{0.764} & \underline{0.866} & \underline{0.883} \\ 
& PointNetVLAD & 0.561 & 0.719 & 0.722 \\ 
& DiSCO & 0.745 & 0.854 & 0.796 \\ 
& EgoNN & 0.663 & 0.797 & 0.850 \\ 
& \textbf{DeepRING (ours)} & \textbf{0.826} & \textbf{0.904} & \textbf{0.901} \\
\bottomrule[1pt]
\end{tabular}
\begin{tablenotes}
        \footnotesize
        \item[*] The compared methods are evaluated on the whole trajectory. For generalization evaluation, the learning-based methods are trained on one dataset and generalized to the other dataset.
      \end{tablenotes}
\end{threeparttable}
\vspace{-0.2cm}
\end{table}

\begin{table}[t]
\renewcommand\arraystretch{1.5}
\centering
\caption{Ablation Study on Diffferent Components}
\label{total ablation}
\begin{threeparttable}
\begin{tabular}{cccccc}
\toprule[1pt]
\textbf{Ablation} & \textbf{Input} & \textbf{Aggregation} & \textbf{Loss} & \textbf{Recall@1} \\ \hline
\multirow{1}{*}{Representation} & PG & DFT & CE & 0.834 \\ \hline
\multirow{3}{*}{Aggregation} & SG & GMP & CE & 0.534 \\ 
& SG & GAP & CE & 0.639 \\ 
& SG & Multi-GAP & CE & 0.743 \\ \hline
\multirow{2}{*}{Loss} & SG & DFT & Triplet & 0.644 \\ 
& SG & 2DFT & Triplet & 0.747 \\ \hline
\multirow{1}{*}{\textbf{Ours}} & SG & DFT & CE & 0.859 \\
\bottomrule[1pt]
\end{tabular}
\begin{tablenotes}
        \footnotesize
        \item[*] PG: Polar Gram, SG: Sinogram, DFT: Discrete Fourier Transform, CE: Cross-Entropy, GMP: Global Max Pooling, GAP: Global Average Pooling, Multi-GAP: Multi-channel Global Average Pooling.
      \end{tablenotes}
\end{threeparttable}
\vspace{-0.5cm}
\end{table}


\textbf{Class Number for one-shot learning:}
With respect to one-shot learning algorithm, we conduct experiments on NCLT dataset to examine the effect of classes/ways number on place recognition performance. As shown in Tab. \ref{way}, the number of classes/ways used in the training stage does not obviously facilitate the final performance. We attribute this to the roto-translation invariance design of our method, which shrinks the intra-class variation. Therefore, the way number shows little influence on the place recognition performance and we take 24-way 1-shot learning algorithm for evaluation in all experiments.


\begin{table}[htbp]
\renewcommand\arraystretch{1.5}
\centering
\caption{Ablation Study on Class Number for Training}
\label{way}
\begin{tabular}{ccccc}
\toprule[1pt]
No. Class & 8 & 16 & 24 & 32 \\ \hline
Recall@1 & 0.834 & 0.855 & \textbf{0.859} & 0.855 \\ 
\bottomrule[1pt]
\end{tabular}
\vspace{-0.4cm}
\end{table}

\section{CONCLUSIONS}
In this paper, we propose a novel framework to learn roto-translation invariant representation for LiDAR based place recognition. Leveraging Radon transform, we convert a LiDAR scan to a rotation equivariant representation SG. Based on circular convolutions and DFT, we exploit rotation equivariant and translation invariant feature maps. Considering place recognition as a one-shot learning problem, we learn the relation between inputs based on correlation distance metric for similarity comparison, which is roto-translation invariant. Thanks to the one-shot learning framework with explicit design of roto-translation invariant representation, our method achieves outstanding performance and recognizes unseen places easily.




\printbibliography

\end{document}